\documentclass[sigconf]{acmart}

\AtBeginDocument{%
  }

\copyrightyear{2022} 
\acmYear{2022} 
\setcopyright{acmcopyright}\acmConference[xx '22]{xxx}
\acmBooktitle{}

\usepackage{color}
\usepackage{amsthm}
\usepackage{booktabs}
\usepackage{algorithm}
\usepackage{algorithmic}
\usepackage{graphicx}
\usepackage{color}
\usepackage{amsmath}
\usepackage{multirow}
\usepackage[misc]{ifsym}
\usepackage{hyperref}
\usepackage{balance}
\begin{document}
\renewcommand\footnotetextcopyrightpermission[1]{}
\title{Delving into the Continuous Domain Adaptation}

\author{Yinsong Xu}
\authornote{Contributed equally.}
\authornote{Work done during an internship at the National Institute of Health Data Science, Peking University.}
\affiliation{%
  \institution{
  School of Artificial Intelligence, Beijing University of Posts and Telecommunications}
  \institution{
  National Institute of Health Data Science, Peking University}
  \state{Beijing}
  \country{China}
}
\email{xuyinsong@bupt.edu.cn}

\author{Zhuqing Jiang}\authornotemark[1]
\affiliation{%
  \institution{
  School of Artificial Intelligence, Beijing University of Posts and Telecommunications}
  \institution{Beijing Key Laboratory of Network System and Network Culture}
  \state{Beijing}
  \country{China}
}
\email{jiangzhuqing@bupt.edu.cn}

\author{Aidong Men}
\affiliation{%
  \institution{
  School of Artificial Intelligence, Beijing University of Posts and Telecommunications}
  \state{Beijing}
  \country{China}
}
\email{menad@bupt.edu.cn}

\author{Yang Liu}
\affiliation{%
  \institution{
  Wangxuan Institute of Computer Technology, Peking University}
  \state{Beijing}
  \country{China}
}
\email{yangliu@pku.edu.cn}

\author{Qingchao Chen}
\authornote{Corresponding author.}
\affiliation{%
  \institution{
  National Institute of Health Data Science, Peking University}
  \state{Beijing}
  \country{China}
}
\email{qingchao.chen@pku.edu.cn}

\renewcommand{\shortauthors}{Yinsong Xu et al.}
\begin{abstract}
Existing domain adaptation methods assume that domain discrepancies are caused by a few discrete attributes and variations, e.g.,  art, real, painting, quickdraw, etc. We argue that this is not realistic as it is implausible to define the real-world datasets using a few discrete attributes. Therefore, we propose to investigate a new problem namely the \textit{\textbf{Continuous Domain Adaptation}} (CDA) through the lens where infinite domains are formed by \textit{continuously} varying attributes. Leveraging knowledge of two labeled source domains and several observed unlabeled target domains data, the objective of CDA is to learn a generalized model for whole data distribution with the continuous attribute. Besides the contributions of formulating a new problem, we also propose a novel approach as a strong CDA baseline. To be specific, \textit{firstly} we propose a novel alternating training strategy to reduce discrepancies among multiple domains meanwhile generalize to unseen target domains. \textit{Secondly,} we propose a continuity constraint when estimating the cross-domain divergence measurement. \textit{Finally,} to decouple the discrepancy from the mini-batch size, we design a domain-specific queue to maintain the global view of the source domain that further boosts the adaptation performances. Our method is proven to achieve the state-of-the-art in CDA problem using extensive experiments. The code is available at \href{https://github.com/SPIresearch/CDA}{https://github.com/SPIresearch/CDA}.
\end{abstract}




\maketitle
\section{Introduction}


\begin{figure}[t]
\centering
\includegraphics[width=0.9\columnwidth]{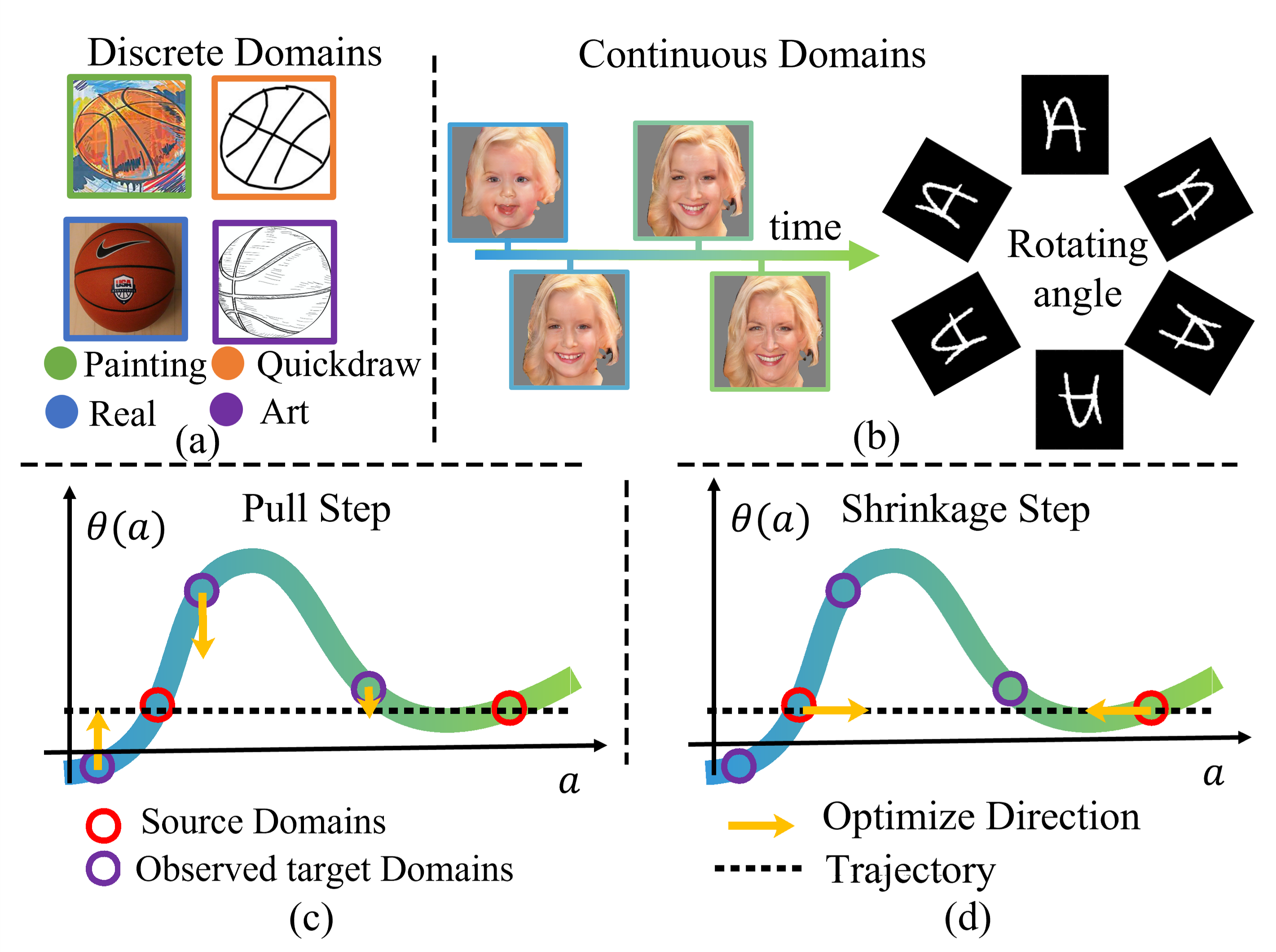} 

\caption{(a)Discrete domains; (b) Continuous domain; (c) Pull step in our method; (d) Shrinkage step in our method. (best view in color, $\theta$ is distribution parameter and $a$ is continuous attribute).}.

\label{fig_teaser}
\end{figure}

Recently, machine learning models trained with large-scale data and well-curated annotations have achieved a steady and significant improvement. However, as the distribution is complex in multimedia content interpretation, models are fragile and may break down under unseen or out-of-distribution (OOD) test domain -- when there exist domain distribution discrepancies between the source (train) and target (test) domain. Given an unlabeled target dataset with domain discrepancies from the source one, it is costly and challenging to annotate enough training data to train a generalized model. The domain adaptation (DA) paradigm aims to minimize the domain discrepancies and successfully transfer the generalized model from the source to the target domain. 

Existing DA methods address the discrepancies among \textit{discrete} domains by learning domain-invariant features. As shown in Figure \ref{fig_teaser}(a), they assume that the domain discrepancies are caused by different discrete concepts in the data collection procedure \cite{peng2019moment}, e.g., art, real, painting, quickdraw, etc. As shown in Table \ref{tab:da_lists}, existing multi-source \cite{zhao2020multi,lin2020multi,zhao2021learning,he2021multi,peng2019moment}, multi-target DA \cite{8970464}, and open compound DA \cite{compounddomainadaptation} tasks assume that the discrepancies are caused by unknown and discrete domain attributes in training. Evolving DA \cite{NEURIPS2020_fd69dbe2} is a continual learning based method that aims to adapt target data that come in an online manner without forgetting. However, practical scenarios are way more complicated, and the previously mentioned assumptions are easily violated. Models are expected to adapt to continuous domains in loads of scenarios, which is more challenging, but there are very few attempts in the literature.

Continuous Indexed DA \cite{wang2020continuously} addresses the adaption in continuous domains, but it requires domain labels for each sample which is unrealistic and restricts the scalability of the approach. In this paper, we consider a more realistic setting of domain adaptation — adapting to continuous domains. Assume that a face recognition system is trained on several labeled photos (the source domains), such as an ID card. When the model is deployed in the real world, the system is expected to recognize targets in the wild, potentially covering long periods (the continuous target domains) such as tracking the suspect or finding the lost child. Another restriction we are confronted with is the domain labels. It is challenging to recognize the shooting time of photos as they are collected from various and uncertain sources. Thus, it is necessary and valuable to learn representations using data collected in a limited period but adapt the models to photos taken at anytime (continuously) in the real world. This setup brings challenges to previous DA methods.

We formulate the above setting as a new setup called continuous DA (CDA): 1)multiple domains are sampled continuously based on an attribute variable, e.g., time and angle, as shown in Figure \ref{fig_teaser}(b). 2)we have access to two arbitrarily sampled labeled \textit{source domains} data and several unlabeled \textit{probe target domains'} data. 3) the \textit{\textbf{task}} is to learn a domain-invariant and discriminate model that generalizes well to domains with the arbitrary attribute value. CDA is different from other existing DA tasks (see Figure \ref{fig_teaser} and Table \ref{tab:da_lists}): \textit{Firstly} in our CDA setup, \textit{infinite} domains are generated based on the continuous attribute values (e.g. time and angle), \textit{rather than} using several discrete attributes in other setup (e.g. art/real/painting style). \textit{Secondly}, domain labels (the attribute values) are not available in the CDA training, ensuring that it is a plausible and real-world problem (but we assume knowing which source domains the source data come from). \textit{Thirdly}, the generalization capability to open domains (target domain data under test are not available in the training stage) is the main evaluation in the CDA. \textit{Finally}, we propose to train the model using only \textit{two} source domains data and annotations, as we think it is good and sufficient enough to model the domain attributes in the CDA problem.

\begin{table}[t]
\small
\centering
\caption{Adaptation tasks using multiple domains, including Multi-Source DA(MSDA), Multi-Target DA(MTDA), Open Compound DA(OCDA), Continuous Indexed DA(CIDA), Evolving DA(EDA) and CDA.} 
\setlength\tabcolsep{0.6pt}
\begin{tabular}{cccccc}%
\hline
DA  & Source  & Target & Domain & Open & Domain  \\
SetUp  & Domain & Domain &Labels  &Domains  & Attribute \\
\hline
MSDA & Multiple & Single & Yes & No & Unknown \\
MTDA & Single & Multiple & Yes & No & Unknown \\
OCDA & Single & Multiple & No & Yes & Unknown \\
CIDA & Multiple & Multiple & Yes & Yes & Discrete \\
EDA & Single & Multiple & No & No & Discrete/Continuous \\
CDA & Two & Infinite & No & Yes & Continuous \\
\hline
\end{tabular}

\label{tab:da_lists}
\end{table}

Through understanding and discussing the characteristics of CDA, we articulate and prosper the challenges as follows: 

\textbf{\textit{Unknown and complex continuity of domain attributes and discrepancies:}} Although the domain attribute is defined smoothly and the value varies continuously, it is computationally challenging to model such continuity in the attribute space using the high-dimension observed data. The complex attributes and their continuities give rise to the challenging and unpredictable domain discrepancies. 

\textbf{\textit{Infinite and non-overlapping target domains under test:}} Although a few target domains data can be sampled to probe the continuous attribute variations, the discrepancy between adjacent target domains is still large and difficult to model. In other words, the probe target domain data available in the training stage are not representative of all the target domain samples drawn in the continuous attribute space. To make things worse, there are infinite domains in the CDA. Therefore it is not guaranteed the model can generalize well to unseen \textit{adjacent} target domain).

\textbf{\textit{Unknown and unpredictable target domain attribute index:}} Predicting the target domain index (attribute) is always challenging based on the finitely sampled probe target domains in the CDA. In addition, we don't assume to access the domain indexes (attribute values) in the CDA.

In this work, we design a novel framework to tackle the CDA challenges.

\textbf{\textit{Firstly}}, to capture the complex domain attributes and improve model generalization to unseen and infinite target domains, we propose a novel \textit{alternating direction training strategy} composed of \textit{Pull(P) and Shrinkage(S) Step}. As shown in Figure \ref{fig_teaser}(c)(d), we model the data distribution parameterized by $\theta(a)$($\theta$ is one dimension for better illustrated and $a$ is the attribute variable) as a trajectory. The P step \textit{firstly} pulls probe target domains on the trajectory path formed by two source domains by reducing the sum of two target-source discrepancies (keep the two source domains fixed); The S step \textit{secondly} shrinks the domain trajectory distance by reducing the discrepancies between two source domains (keep the target domains fixed). The intuition behind is: alternating the P and S step progressively, the multi-domain attribute geometry formed by source and probe target domains is preserved. In addition, the strategy guarantees the generalized solutions to unseen target ones (see ablation study results in Table \ref{analy}). 

\textbf{\textit{Secondly}}, it is challenging to estimate the complete domain attribute geometry and pull all unseen target domains onto it \textit{using only two source domains}. To tackle this problem, we propose a novel continuity constraint when estimating the cross-domain divergence measurement. In addition, we provide remarks and implications of the regularized constraint. This helps pull unseen target domains to the domain attribute trajectory.

\textbf{\textit{Finally}}, in the CDA setup, as the proposed domain discrepancy needs to cover and pull infinite domains to the trajectory in the stochastic manner, we propose a novel implementation to maintain the global view of source domain statistics using a queue. It is able to improve the source model's ability to adapt by preserving more complete source information than mini-batches.  

Our contributions are summarized as follows:
\begin{itemize}
\item We propose a new and realistic problem namely the Continuous Domain Adaptation. 
\item We propose a novel alternating direction training strategy and a domain-continuity regularizer to reduce multi-domain discrepancies meanwhile maintaining the geometry of the domain attributes. In addition, a novel queue based implementation is designed to estimate the global source domain statistics.
\item Our analysis provides insight and our method achieves SOTA results. 
\end{itemize}

\section{Related Works}

\textbf{Single-source domain adaptation.} Existing methods explicitly measure and reduce the single source-target cross-domain discrepancy, including Maximum Mean Discrepancy(MMD) \cite{tolstikhin2016minimax}, $\mathcal{H} \Delta \mathcal{H}$-divergence \cite{ben2010theory}, Maximum Classifier Discrepancy \cite{saito2018maximum}, Margin Disparity Discrepancy (MDD) \cite{zhang2019bridging}, and Optimal Transport\cite{ijcai2021-394}, Source-free UDA\cite{10.1145/3474085.3475487}, Contrastive Learning \cite{10.1145/3474085.3475496}, and the adversarial learning methods \cite{10.1145/3343031.3351070, Chen_2018_CVPR,Li21BCDM,Liang_2021_CVPR,zhong2021does, 10.1145/3394171.3413701, 10.1145/3394171.3413516}.
\textbf{\textit{However}}, as there are infinite target domains in CDA, most of these methods show degraded performances or challenging to adapt in CDA setting.

\noindent\textbf{Multi-domain adaptation.} Several works address the multi-domain adaptation problem listed in Table \ref{tab:da_lists}
\cite{zhao2020multi,compounddomainadaptation,jin2020minimum,he2021multi} but they only consider the domains sampled on \textit{discrete} attribute variable. The most similar work to ours is continuous-indexed domain adaptation method \cite{wang2020continuously} however, they assume the domain indexes are available in the training. \textbf{\textit{However}}, in the CDA and our method, target domain indexes are not available and we assume only two source domains available in the training, which is more realistic setup. 

\noindent\textbf{Domain generalization(DG).} DG methods \cite{li2018domain,dg_mmld,li2018learning,zhao2021learning, 10.1145/3394171.3413852, 10.1145/3474085.3475311} aim to train a model with labeled data that can generalize to any unseen target domain. Similar to DG setup, the CDA setup assumes to access some target domain's data but not all target domains. \textbf{\textit{Differently}}, our domain attribute varies in the continuous manner and the CDA uses two source domains. 

\noindent\textbf{Continual learning based domain adaptation.} 
Existing domain adaptation methods based on CL focus on ``continual'' adaptation however we adapt for ``continuous'' domains. 
\cite{bobu2018adapting,lao2020continuous,NEURIPS2020_fd69dbe2,su2020gradient,mancini2019adagraph} focus on learning target tasks online and they test on seen target domains, \textbf{\textit{however}} our CDA problem assumes that there are \textit{infinite, continuously changing and unseen} target domains under test and with unknown domain attribute. 

\section{Continuous Domain Adaptation}
\begin{figure}[t]
\centering
\includegraphics[width=0.95\columnwidth]{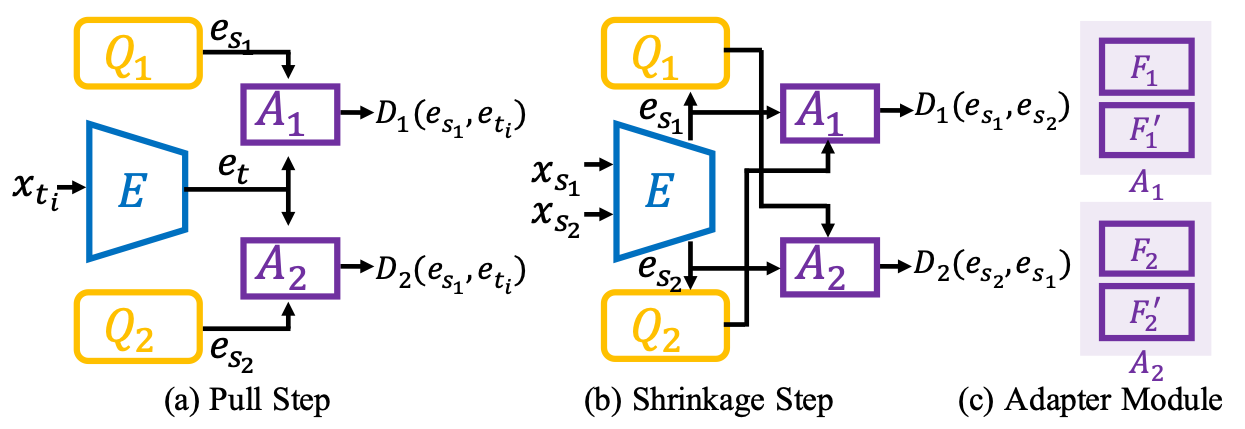}

\caption{Overall framework. We separate the network into three parts: a joint encoder ($E$), adapter modules ($A_1$,$A_2$) and queues ($Q_1$,$Q_2$). (a)Minimize the sum of two target-source discrepancies in Pull Step. (b)Minimize the discrepancy between two source domains in Shrinkage Step. (c)Each adapter consists of a content classifiers $F$ and an auxiliary classifiers $F'$.}
\label{framework}

\end{figure}

\begin{figure*}[t]
\centering
\includegraphics[width=0.9\textwidth]{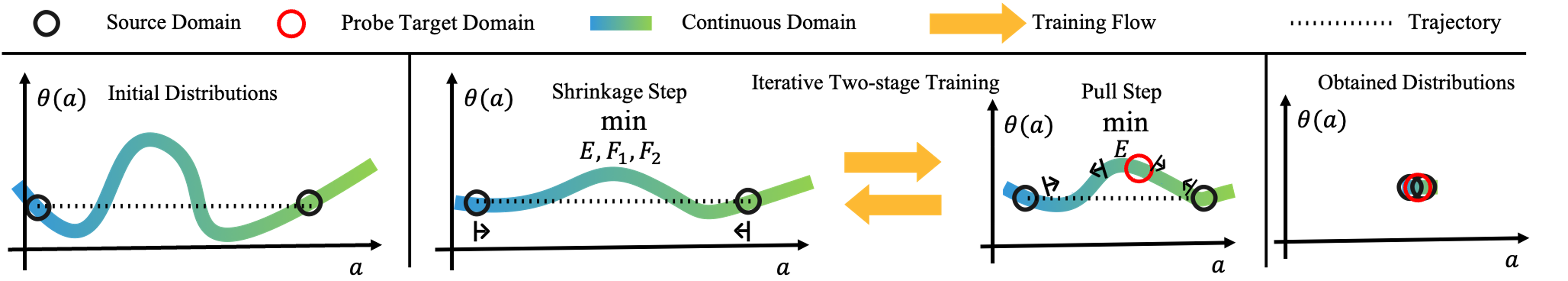} %

\caption{Illustration of the domain distribution geometry with the overview of our method. First, the trajectory formed by two source domains is shown in initial distribution (left sub-figure). Second, in \textit{Pull Step}, target domains are pulled close to the trajectory. In \textit{Shrinkage Step}, the trajectory distance shrinks. Our method optimize two steps in the alternating manner to reduce the multi-domain discrepancy in CDA progressively.}
\label{trajectory_framework}
\end{figure*}




\textbf{Problem Formulation:} In our CDA setup, the data distribution is given by $P_{\theta(a)}(x,y)$ ($P_{\theta(a)}$ for abbreviation), parameterized by $\theta(a)$, and $a$ is an unavailable continuous attribute in the real world scenarios (e.g., time, rotation). We are provided with \textit{two} labeled source domains $\{S_i(x,y)\}_{i=1}^{2}$ from $P_{\theta(a_i)}$, and a set of unlabeled probe target domains $\{T_j(x)\}_{j=1}^{N_T}$ from $P_{\theta(a_j)}$. The goal of CDA is to learn a generalized model that achieves good performance on $P_\theta(a)$ with arbitrary $a$, leveraging the knowledge of two source domains (data and annotations) and limited number of target domains' data. 

\textbf{How fragile is DA method in CDA problem?} CDA is analogous to a practical setting where we are given a dataset in the real-world application with a continuously varying attribute. Furthermore, such attribute variations, although sometimes tiny, will destroy the model developed using partially annotated data from the dataset (see Benchmark Analysis in Sec 5.1). 

\textbf{Why two source domains?} In the CDA task, all domains are sampled based on the continuous attribute, and the domain variations are mainly caused by the attribute variations. Thus, we are given two labeled source domains and a limited number of target domains to exploit the latent attribute and its variation. Note that the test data also comprises unseen domains. 

\section{Method}
\subsection{Motivation}
The main objective of CDA is to learn a generalized model tackling discrepancies among all domains sampled on a continuous attribute. To handle the statistics of unseen domains, we model each distribution parameter $\theta$ as a point in the high-dimension space, as shown in figure \ref{trajectory_framework}. Thus, connecting all values of the attribute $a$, we can obtain a trajectory $\theta(a)$ in the space. In this way, infinite domains' data distribution is represented as a trajectory. Our strategy is to study the geometry of the continuous attribute and use it as an inductive bias in the modeling. Inspired by this principle, we design the alternating direction training strategy, including two steps: Pull and Shrinkage. As shown in figure \ref{trajectory_framework}, in \textit{Pull Step}, we regard the trajectory formed by the two source domains as the inductive bias and pull target domains to the trajectory so that unseen target domains have a higher chance to be close to the trajectory as well. In \textit{Shrinkage Step}, we progressively shrink and shorten the trajectory. Alternating the two steps in training, the proposed method is proved to generalize to unseen target domains.

As unseen target data is infeasible in training, we approximate it with disturbance on probe target data through which the source-unseen target discrepancy can be estimated by source-probe target discrepancy. When the disturbance is limited to $0$, gradient plenty can be derived as a constraint of the discrepancy. Then, the discrepancy between the unseen target and source domain is minimized by the constraint term. In addition, to obtain global discrepancy by perceiving more source information than a mini-bacth, we decouple the discrepancy from the mini-batch size by utilizing queues.

\subsection{Overall Framework}
The overall architecture is shown in figure \ref{framework}. Specifically, the framework comprises a joint feature encoder $E$, which extracts the feature $e=E(x)$ of input data $x$, and two adapter modules $A_1$ and $A_2$, each of which consists of two paired classifiers. 
The reason of adopting two adapter modules rather than one is that, one adapter \textit{fails to simultaneously fit} two source domains during early training. As a result, it estimates biased cross-domain discrepancy which leads to fragile optimized results. Thus, we design two pairs of classifiers ($F_1$,$F_1'$) and ($F_2$,$F_2'$) to classify source samples and tackle two discrepancy measurements ($\mathcal{D}_1$,$\mathcal{D}_2$) for two source domains respectively. 

To improve model generalization to unseen and infinite target domains, \textit{a novel alternating direction strategy} is proposed to progressively reduce multiple domains' discrepancies, as shown in Figure \ref{trajectory_framework}. The alternating direction training strategy is composed of two alternating steps. To be specific, in \textit{Pull Step}, we compute discrepancies between probe target domain $T_i$ and two sources domains $S_1$, $S_2$. Then we reduce the sum of them to pull all target domains close to the trajectory formed by two source domains. In \textit{Shrinkage Step,} the distance of the trajectory is shortened by reducing the discrepancy between two source domains. Alternating the two steps, the training strategy enables more stable and robust optimization. 

The final objective of CDA is to improve the model generalization on the whole data distribution. Nevertheless, the above strategy only reduces discrepancies between source and the seen probe target domains, without considering the unseen target ones. To overcome it, \textit{a Continuity Constraint} is proposed to regularize the continuous geometry of domain attributes by regularizing the gradient penalty in Pull Step. It is useful because it implicitly constrains that pulling probe target domains to source ones tends to pull the unseen target ones.


As there are infinite target domains and our method has to pull finite target domains to the source ones in the stochastic manner, it is challenging to maintain the global semantic features of the source domain using mini-batch methods. Therefore, a novel implementation is adopted to maintain the global view of two source domains using queues $Q_1$ and $Q_2$, preserving and updating the complete and up-to-date source domain information. The queue has the following advantages over the previous mini-batch form: 1) the queue size is flexible to set and can be much larger than a mini-batch size; 2) the queue always maintains the newest source domain features.

\subsection{Alternating Direction Strategy for Multi-Domain Discrepancy Reduction}

\textbf{Multi-Domain Discrepancy.} The hypothesis-induced discrepancies \cite{ben2010theory, zhang2019bridging} require taking supremum over hypothesis space to measure the discrepancy of two domains. In this work, we extend it into multi-domains. Specifically, we propose two pairs of classifiers. Content classifiers $F_1$ and $F_2$ are used to classify content labels in each source domain, while auxiliary classifiers $F_1'$ and $F_2'$ are used to estimate the discrepancies (together with content classifiers) between each source domain and the other target domains for the previously mentioned alternating training strategy. The discrepancy of domain $P$ and $Q$ is formed using the supremum of prediction differences between a pair of classifiers $F_j,F'_j (j=1,2)$ as follows: 
\begin{align}
\label{sup}
\mathcal{D}_j(e_{p},e_{q}) = \sup_{F'_j}\big[-\mathbb{E}_{e_{p}\sim P}\log[\sigma_{h_{F_j}(e_p)}\circ F'_j(e_p)] \notag \\
                    - \mathbb{E}_{e_q\sim Q}\log[1-\sigma_{h_{F_j}(e_q)} \circ F'_j(e_q)]\big],
\end{align}
where $\circ$ denotes function composition, $h_F$ is a labeling function: $h_F(e)=\arg\max_{y}p(y|x)$, where $p$ is the softmax probabilities predicted by $F$, and $\sigma$ is the softmax function, $\sigma_j(z)=\exp(z_j)/\sum_i\exp(z_i)$.

\textbf{Advantages of using two adapters and discrepancies.} Due to using two source domains, we adopt $F_i$ and ${F_i}^{'}$, $i \in \{1,2\}$, to estimate two separate discrepancies between some target and each ($i^{th}$) source domain. It is hard for one classifier to fit two source domains at the beginning of training because they lie in different trajectory positions. Our discrepancy measurement is based on classifiers, it is more accurate to estimate two discrepancies separately. 

\textbf{Pull Step.} Feeding samples from $S_1,S_2$ and $T_i$ to the network, we train the encoder $E$ to \textit{pull} the target domains on the trajectory path formed by the two source domains. It is achieved by reducing the sum of $\mathcal{D}_1(e_{s_1},e_{t_i})$ and $\mathcal{D}_2(e_{s_2},e_{t_i})$, which reaches the minimum when $T_i$ is on the trajectory. Networks can be trained based on the following objective, $\mathcal{L}_p$:
\begin{align}
\label{objp}
\min_{E} \sum_{i=1}^{N_T}\mathcal{D}_{1}(e_{s_1},e_{t_i}) + \mathcal{D}_{2}(e_{s_2},e_{t_i}). 
\end{align}

\textbf{Shrinkage Step.} After pulling target domains to the trajectory in the Pull Step, source-only samples are utilized to optimize the encoder $E$ and two classifiers $F_1$, $F_2$ to \textit{shrinkage} the trajectory distance by reducing the following discrepancy between two source domains, denoted as $\mathcal{L}_s$.
\begin{align}
\min_{E} \mathcal{D}_1(e_{s_1},e_{s_2})+ \mathcal{D}_2(e_{s_2},e_{s_1}).
\label{objs}
\end{align}

As the source labels are available, the cross-entropy loss $\mathcal{E}(x,y)$ is adopted to train the networks $E,F_1,F_2$ as follow, denoted as $\mathcal{L}_{ce}$: 
\begin{align}
\min_{E,F_1,F_2} \mathcal{E}(x_{s_1},y_{s_1}) + \mathcal{E}(x_{s_2},y_{s_2}).
\end{align}

\textbf{Remarks.} One may argue that we should train two steps together in a unified framework. Although it is practicable, we argue that it will pull both target and source domains together and the geometry of the trajectory will change drastically. As a result, the continuity will not be constrained effectively, which is harmful for generalization on unseen target domains. The comparison result is shown in Sec 5.3.

\subsection{Theoretical Insights and Explanations}
As we take supremum to define the discrepancy in Eq \eqref{sup}, we first show the optimum of $F'_i$ that reaches the supremum and the formulation of loss function Eq \eqref{objp} and Eq \eqref{objs} in two P Step and S Step, respectively. Denoting $P_{s_j}$ and $P_{t_i}$ as the distributions of two domains' features $e_{s_j}$ and $e_{t_i}$ encoded by $E$, we derived the optimal values of $\sigma_{h_{F(e)}}\circ F'(e)$ in P and S Steps as follows:

\noindent In Pull Step: 
\begin{align}
\label{op1}
 \sigma_{h_{F_j(e)}}\circ F'_j(e) = \frac{ P_{s_j}}{P_{s_j} + P_{t_i}},  j=1,2.
\end{align}

\noindent In Shrinkage Step:
\begin{align}
\label{op2}
 \sigma_{h_F(e)}\circ F'_1(e) = \frac{ P_{s_1}}{P_{s_1} + P_{s_2}}, \sigma_{h_F(e)}\circ F'_2(e) = \frac{ P_{s_2}}{P_{s_2} + P_{s_1}}.
\end{align}

Now we analyze the convergence of the proposed alternating learning algorithm where the minimization of Eq.\eqref{objp} and Eq.\eqref{objs} will align source and target domains. Substituting Eq.\eqref{op1} and Eq.\eqref{op2} in Eq.\eqref{objp} and Eq.\eqref{objs} respectively, the objective can be derived into:
\begin{equation} 
\min_{E} \left\{
\begin{aligned}
& \sum_{i=1}^{N_T} \text{JS}(P_{s_1}||P_{t_i}) + \text{JS}(P_{s_2}||P_{t_i}), &\text{P Step}\\
& 2\text{JS}(P_{s_1}||P_{s_2}).  &\text{S Step} \label{objpp}\\
\end{aligned}
\right.
\end{equation}
\noindent It shows that the sum of JS-divergence between target domains and two source domains is minimized
in P Step, while the JS-divergence between two source domains is minimized in S Step.

\textbf{Global optimum.} Since the JS-divergence between two distributions is always non-negative and zero only when they are equal. Thus, the solution of P Step in Eq \eqref{objpp} is $p_{s_1}=p_{t_i}$ and $p_{s_2}=p_{t_i}$, while that of S Step is $p_{s_1}=p_{s_2}$. Two steps run alternatively and reach global minimum when $p_{s_1}=p_{s_2}=p_{t_i}$ i.e., the domains are aligned perfectly and the trajectory becomes a point as shown in the right part of Figure \ref{trajectory_framework}.






\subsection{Continuity Constraint via Gradient Plenty}
The above strategy is not guaranteed to pull the unseen target domains close to the trajectory. To address it, we propose to add the continuity constraint in the optimization. Given the probe and the neighborhood unseen target domain with attribute $a$ and $a+\Delta a$ respectively, their encoder features can be represented as $e_{a}$ and $e_{a+\Delta a}$. Though estimating the discrepancy $\mathcal{D}_i(e_{s_i},e_{t(a+\Delta a)})$ accurately is intractable, we can find a constant $\eta$ that:
\begin{align}
\mathcal{D}_i(e_{s_i},e_{t(a+\Delta a)}) \leq \eta \Delta a + \mathcal{D}_i(e_{s_i},e_{t(a)})).
\label{gp}
\end{align}
As the domain variations are mainly caused by the attribute variations, when $\Delta a \to 0$, $e_t{(a+\Delta a)} = e_t(a)+\beta \Delta a $, where $\beta$ is another constant. Therefore, Eq.\eqref{gp} can be reformulated as:
\begin{align}
\left \| \nabla_{e_{t}} \mathcal{D}_i(e_{s_i},e_{t}) \right \| \leq \eta/\left \|\beta\right \|. \end{align}
Empirically we set ${\eta}/{\left \|\beta\right \|}=1$ and implement it as the gradient penalty in \textit{Pull Step}, denoted as $\mathcal{L}_{gp}$:
\begin{align}
\min_{E} (\left \| \nabla_{e_{t}} \mathcal{D}_i(e_{s_i},e_{t}) \right \|-1)^2.
\end{align}
In this way, when we minimize the discrepancy of between the source and probe target domains, $\mathcal{D}_i(e_{s_i},e_{t})$, the discrepancy of between the source and unseen target domains , $\mathcal{D}_i(e_{s_i},e_{t(a+\Delta a)})$, is minimized implicitly. 
\subsection{Global Domain Discrepancy Measurement using Queue Implementation}
The fundamental problem of the alternating direction strategy and the CDA task is to robustly estimate the domain shifts in the stochastic manner. Existing discrepancy computation performance heavily relies on mini-batch size, especially in CDA task where there are infinite target domains under test and the discrepancy is biased.  
To decouple the discrepancy from the mini-batch size, we introduce queue to obtain more global and complete cross-domain discrepancy. Concretely, in \textit{Shrinkage Step}, the samples in two source domains are progressively replaced in two queues, denoted as $Q_1$ and $Q_2$. In \textit{Pull Step}, we compute the discrepancy between the target domain and two source domains. 

\subsection{Overall Optimization}
The overall optimization is as follows:
\begin{align}
     & \text{Pull Step: } &\min_{E} \mathcal{L}_p + \mathcal{L}_{gp}, \\
     & \text{Shrinkage Step: } &\min_{E,F_1,F_2} \mathcal{L}_s + \mathcal{L}_{ce} .
\end{align}










\section{Experiments}
We first introduce a new CONtinuous Domain Adaptation (CONDA) benchmark to investigate the CDA problem and evaluate the proposed method. Then the detailed analysis of the benchmark is performed and different splits are selected for efficient evaluations. Finally we compare with other methods for extensive evaluations.
\begin{figure}[t]
\centering
\includegraphics[width=0.7\columnwidth]{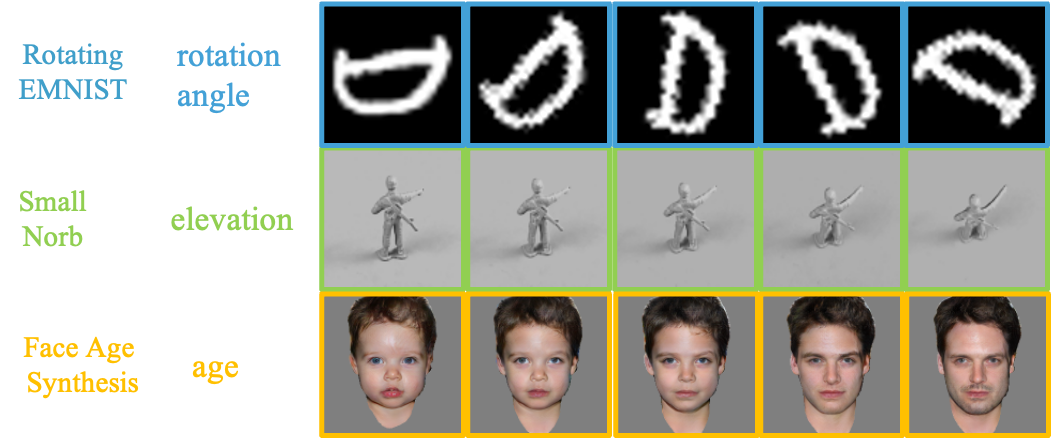} 
\caption{Adopted datasets: Rotating EMNIST, SmallNorb and Face Age Synthesis. Three different attributes are selected in those datasets: rotation angle, elevation and age.}
\label{sample}
\end{figure}
\begin{figure}[t]
\centering
\includegraphics[width=0.9\columnwidth]{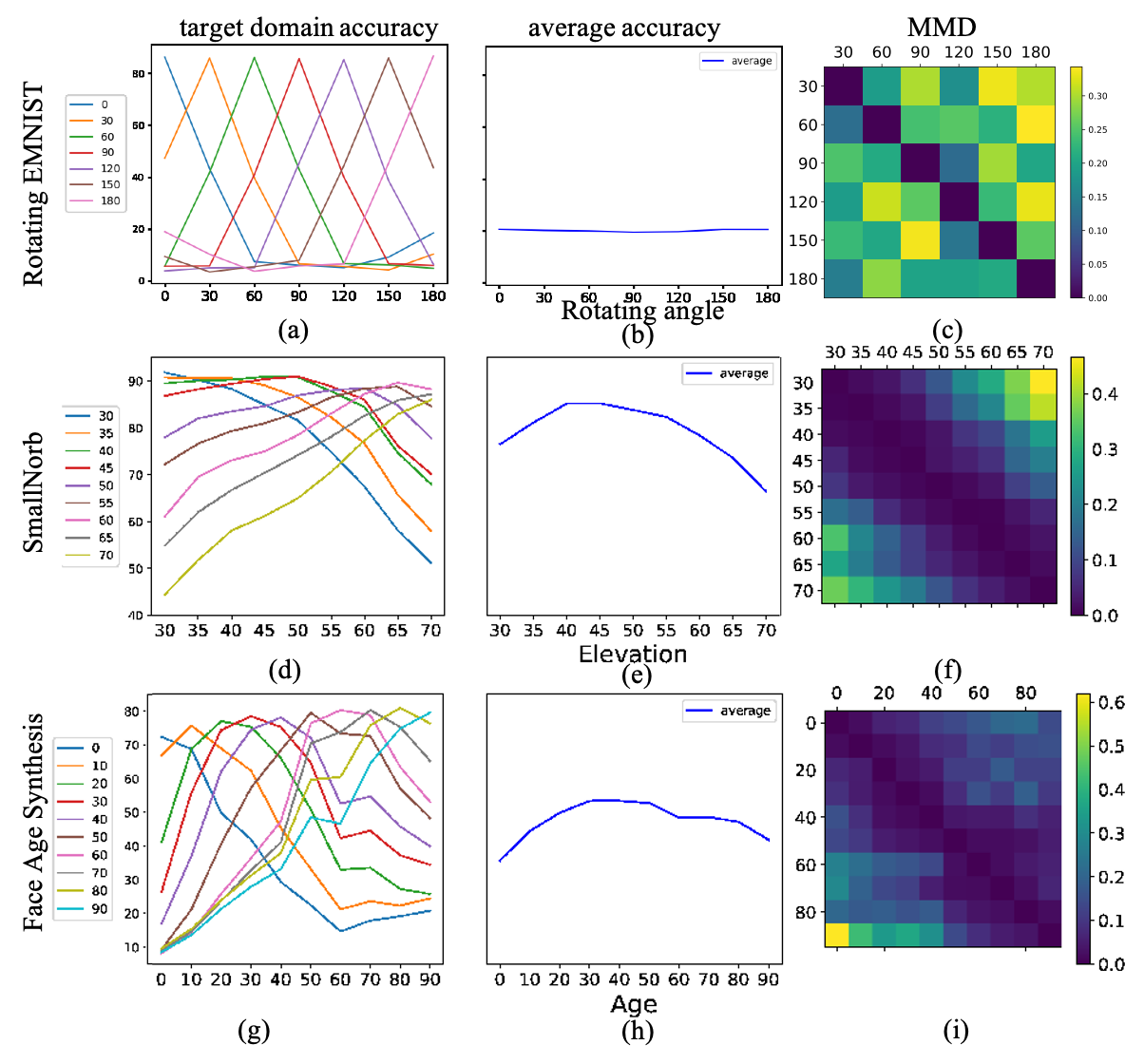}

\caption{Benchmark analysis including target domain accuracy, average accuracy and MMD between source and target domain features. }
\label{data}

\end{figure}
\begin{figure}[t]
\centering
\includegraphics[width=0.95\columnwidth]{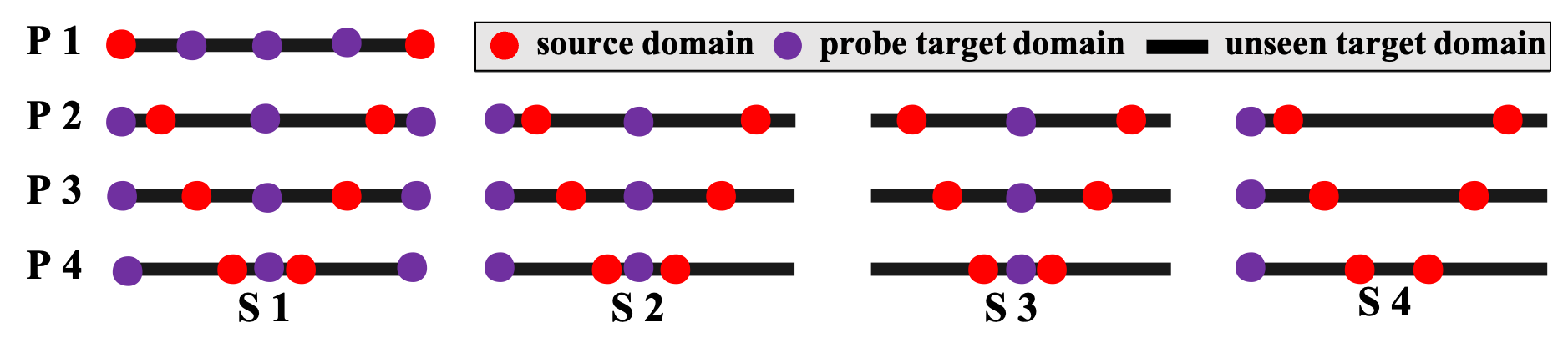}

\caption{We design 13 splits for CDA according to the source domain locations in the attribute space and the parts to sample probe target domains.$P$ and $S$ denote positions of source and probe target domains, respectively.}

\label{splits}
\end{figure}
\subsection{Continuous Domain Adaptation Benchmark}
\label{sec:benchmark}
\noindent\textbf{Datasets}: The CONDA benchmark consists of in total 474,530 images with full annotations, sourced from three datasets including EMNIST \cite{cohen_afshar_tapson_schaik_2017}, SmallNorb \cite{lecun2004learning}, and our synthesized dataset Face Age Synthesis based on works in \cite{liu2015faceattributes} and \cite{orel2020lifespan}. Figure \ref{sample} shows multiple domains based on a continuously varied attribute. Three different attributes are selected including image rotation angles, the elevation camera views, and the age of the human identities.

\textbf{Rotating EMNIST}\cite{cohen_afshar_tapson_schaik_2017} is an extension of MNIST including six different splits including 131,600 characters of 47 classes. Images are rotated by $0^{\circ}$ to $180^{\circ}$. \textit{We adopt this dataset to study continuous domain variations caused by rotation angles of 2D images.}

\textbf{SmallNorb}\cite{lecun2004learning} contains 48,600 images of 50 toys from 5 generic categories: four-legged animals, human figures, airplanes, trucks, and cars. The toys were imaged under 9 elevations (30 to 70 degrees every 5 degrees). \textit{We adopt this dataset to study continuous domain variations caused by elevation camera views.}

\textbf{Face Age Synthesis.} Age is a common and continuous attribute that causes humans' appearance variations, without existing dataset to support, we propose to use Lifespan Age Transformation Synthesis \cite{orel2020lifespan} to synthesize a full span of 0–90 ages with the interval of 10. 294,330 images of 1000 identity are selected from CelebA\cite{liu2015faceattributes} under synthesis based on the top 1,000 identities FID \cite{xu2018empirical}.\textit{We adopt this dataset to study continuous domain variations caused by age of human identities.}

\begin{table*}[t]

\small
\centering
\caption{Comparision Results (\%).}

\begin{tabular}{llrrrrrrrrrrrrrr}
\hline
\multirow{2}{*}{Dataset}& \multirow{2}{*}{Method}&  P 1& \multicolumn{4}{c}{P 2} & \multicolumn{4}{c}{P 3} & \multicolumn{4}{c}{P 4} &\multirow{2}{*}{Avg}\\
  && S 1&S 1   &S 2     &S 3     &S 4    &S 1     &S 2     &S 3     &S 4    &S 1     &S 2     &S 3     &S 4 \\
\hline  
 \multirow{7}{*}{Rotating EMNIST} &SO &   26.8& 41.7   &41.7     &41.7     & 41.7    &51.9    &51.9     &51.9     &51.9     &44.7    &44.7     &44.7     & 44.7     &44.6  \\
&CIDA&  31.1& 44.7   &45.3    &44.7     &44.1     &57.2    &57.1     &56.2 &56.4  & 49.0    &48.7    &\textbf{48.4} &48.1    &48.5   \\
&BCDM&  33.3&   49.7    &46.6     &50.7    &30.3     &60.0     &60.2     &58.4  &47.8  &45.1  &45.9    &42.2  &48.7  &47.6       \\
&ATDOC& 30.6& 44.5    &43.6    &44.5     &42.5    &60.9     &59.7     &\textbf{61.6}        &\textbf{58.2}     &47.9     &48.4   &47.4   &50.4   &49.2  \\
&StableNet& 26.6 &41.5&41.5&41.5&41.5  &51.9& 51.9& 51.9& 51.9 &44.3&44.3&44.3&44.3 &44.4\\
&MCC & 29.2 &34.8 &34.7 &35.4 &36.2 &40.2 &39.9&40.9 &43.6 &39.9 &38.9 &39.0 &40.6 &37.9\\
\cline{2-16}  
&Ours&  \textbf{34.2} &\textbf{51.1} &\textbf{50.6}     &\textbf{52.1}     &\textbf{45.6}   &\textbf{61.7}      &\textbf{60.8}     &59.4     &55.0    &\textbf{52.3}     &\textbf{51.5}     &48.3   &\textbf{51.6}  &\textbf{51.8}\\
\hline 
\\
\hline 
\multirow{7}{*}{SmallNorb} &SO &85.1 &88.8 &88.8 &88.8 &88.8 & 86.8 &86.8 &86.8 &86.8 &86.5&86.5&86.5&86.5&87.1\\
&CIDA &82.4 &85.5 & 85.9 & 86.8 &85.2 & 85.5 &85.1 &85.7 &85.0 &84.8 &85.2& 84.1& 84.0 &85.0\\
&BCDM &87.4 &78.6&83.0&80.5 &83.5 &77.1 & 80.8 & 79.2 &74.2 &77.4 &82.2 &78.0&74.4&79.7\\
&ATDOC &80.0 &84.4 &87.3 &88.1 &87.9 &83.7 &86.5 &86.7 &86.3 &82.6 &86.5 &86.7 &85.6&85.6\\
&StableNet&80.1&87.0&87.0&87.0&87.0&85.7&85.7 &85.7 &85.7 &83.5 &83.5 &83.5 &83.2 &85.0  \\
&MCC &86.2 &86.3 &86.2 &83.3 &80.1 &86.8 &86.9 &81.7 &82.9 &83.4 &83.7 &75.3 &81.6 &83.4 \\
\cline{2-16}    
&Ours &\textbf{91.5} &\textbf{90.5} &\textbf{90.3} &\textbf{89.6} &\textbf{89.8} &\textbf{87.6} &\textbf{89.5} &\textbf{87.4} &\textbf{89.7} &\textbf{87.8} &\textbf{89.4} &\textbf{87.8} &\textbf{90.1} &\textbf{89.3}\\
\hline
\\
\hline 
 \multirow{7}{*}{Face Age Synthesis} & SO &71.6 &74.1 &74.1 &74.1 &74.1 &71.7 &71.7 &71.7 &71.7 &70.6 &70.6 &70.6 &70.6 &72.1 \\
&CIDA &29.4 &58.6 &53.4 &44.5 &52.2 &53.4 &53.0 &47.2 &49.1 & 55.0 &57.6 &48.0 &53.0 &50.3\\
&BCDM &78.8&76.2 &54.4 &47.3 &54.0 &73.0 &64.4&49.8&47.5&71.9&76.1&65.5 &62.4&63.2\\
&ATDOC &30.8 &49.1 &48.9 &27.2 &48.5 &24.5 &31.3 &25.5 &45.4 &31.9 &41.5 &34.3 &44.6&37.2\\
&StableNet& 69.6 &78.1&78.1&78.1&78.1 &76.6 &76.6 &76.6 &76.6 &74.0&74.0&74.0&74.0 &75.7\\
&MCC &30.4 &55.6&55.9&50.2&57.9 &48.7 &52.8 &56.0 &54.5 &52.6 &58.7 &55.8 &57.9 &52.8\\
\cline{2-16}  
 &Ours &\textbf{79.0}&\textbf{81.8}&\textbf{79.4}&\textbf{79.1}& \textbf{78.2}&\textbf{79.9}&\textbf{78.6}&\textbf{76.0}&\textbf{76.9} &\textbf{79.6}&\textbf{79.4}&\textbf{74.4}& \textbf{77.1} &\textbf{78.4}\\
\hline 
\end{tabular}
\label{result}
\end{table*}
\textbf{Benchmark Analysis}: We conduct an in-depth analysis of the cross-domain statistics in CONDA benchmark in Figure \ref{data}. Three results are reported including the 1) target domain performances using different source-only models; 2) average target domain performances of source-only models; 3) MMD \cite{tolstikhin2016minimax} between source and target domain features using pre-trained source-only model. 

From the Figure \ref{data} (a)(d)(g), we made two observations: \textit{first}, the performance of each source domain model (a plot using a specific color) reaches the peak at the source attribute but successively drops when the attribute keeps moving far away from the source one. This suggests that the domain shift increases with the increasing attribute differences. \textit{Second}, SmallNorb and Face Age Synthesis exhibit the similar smooth performance patterns among all target domains, however, the performance curve in Rotating EMNIST has a very steep peak point on the source attribute and performances drop very quickly for other attributes/domains. This suggests that the current deep network architecture is not suitable to tackle rotation variations and the domain gaps in this dataset seem the largest.
It confirms the fact that deep architectures cannot tackle the domain shift in CDA. 

In the Figure \ref{data} (b)(e)(h), SmallNorb and Face Age Synthesis exhibit the similar average performances patterns among all target domains, where the performance first rises and then falls. Instead, the performance curve in EMNIST is stable among different attributes. It is reasonable because deep networks are not able to handle rotation variations.

Finally, it is observed in Figure \ref{data} (c)(f)(i) that the closer to the diagonal, the darker the point color is. The phenomenon is strongly correlated to the accuracy curve. It provides the insight that the choice of samples to annotate has an impact on the performance when the annotation budget is limited. 

\textbf{Splits.} To perform comprehensive evaluation, we design 13 splits, with the attempt to cover \textit{all combinations of choices of source, probe target and target test domains in the evaluation.} The source domain locations in the attribute space are denoted as P1-P4, while four settings of the segment to randomly sample the probe target denotes as S1-S4, as shown in Figure \ref{splits}.

\subsection{Results and Comparisons in CDA}
\textbf{Implementation.} We perform extensive evaluations on three datasets. For Rotating EMNIST and smallNorb, a four-layer CNN encoder $E$ and a two-layer MLP for each classifier are adopted. For Face Age Synthesis, we use pre-trained ResNet-50 from ImageNet as the encoder. We adopt Adam with learning rate $1 \times 10^{-4}$. See the appendix for more details. The average classification accuracy of all target domains is used in the evaluation. The results of 13 splits on Rotating EMNIST, SmallNorb and Face Age Synthesis are shown in Table \ref{result}. 

\textbf{Comparisons.} In each split, performances of our method and six existing comparison methods are reported. As it is the first time to address the problem of CDA, we \textit{reproduce} existing methods in different tasks. As To summarize, these comparison methods comprise source only(SO) model, \textit{Continuously Indexed DA}(CIDA\cite{wang2020continuously}), \textit{Single Source DA}(BCDM \cite{Li21BCDM}, ATDOC\cite{Liang_2021_CVPR}), \textit{Multi Source/Target DA}(MCC\cite{jin2020minimum}) and \textit{Domain Generalization}(StableNet\cite{zhang2021deep}). 

To summarize the main observations: (1) our method outperforms all other comparison methods in all splits and datasets. It demonstrates that our methods can effectively handle the continuous domain shifts using two source domains. (2) It is worth noting that some methods (i.e., CIDA, ATDOC) perform worse than SO performance. 
\subsection{Ablation Study}
\begin{table}[t]
\small
\centering
\caption{Analysis Results (\%) in SmallNorb Dataset.}
\begin{tabular}{lrrrrrrrr}
\hline
Split &\multicolumn{8}{c}{Variant} \\ 
P2 &V1 &V2 &V3 &V4 &V5 &V6 &V7 &Ours\\
\hline
S1& 87.2& 87.6& 86.6& 86.7& 85.1& 86.5& 89.1& \textbf{90.5}\\
S2& 87.1& 87.5& 88.4& 88.5& 85.4& 89.7& 89.1&  \textbf{90.3}\\
S3& 87.4& 86.4& 88.0& 86.0& 82.5& 88.9& 88.3& \textbf{89.6}\\
S4& 88.8& 87.1& 87.3& 87.0& 87.0& 89.0& 88.0& \textbf{89.8}\\
\hline
\end{tabular}
\label{analy}
\end{table}

In this section, we perform ablation studies on all splits to investigate the effectiveness of the proposed three main contributions: discrepancy measure, two-stage strategy and feature queues. Results are shown in Table \ref{analy}.

We compare the following variant methods and analyze their results. \textbf{V1:} We adopt three binary domain discriminators to encourage domain confusions between ($S_1, S_2$), ($S_2, T$) and ($S_1, T$). \textbf{V2:} We merge P and S Steps into a single step and minimize three pairs of discrepancies at the same time, without fixing $S_1$ and $S_2$ in P Step.\textbf{V3:} We merge P and S Steps into a single step without fixing $S_1$ in P Step.\textbf{V4:} We merge P and S Steps into a single step without fixing $S_2$ in P Step.\textbf{V5:} We remove the S Step but keeping other components in our method. \textbf{V6:} We remove the queue implementation. \textbf{V7:} We remove the continuity constraint.

Compared with our method, the adaptation performances of all variants drastically decrease, and Variant 1 achieves the worst performance. Variant 1 is equivalent to reducing three JS divergence losses. The proposed method outperforms this variant, demonstrating that directly minimizing pair-wise discrepancies fails to generalize to unseen target domain in the continuous domain space. In addition, disjoint domains in continuous domain space leads to unreliable training of the encoder as discussed in \cite{arjovsky2017towards}. From comparison with variants 2-5, merging stages and unfixing source domains in P Step will cause ambiguous optimization direction which is harmful for adaptation. For example, if we unfix the source domain, the geodesic path is dynamic and the target domains are hard to converge. Comparing with variant 6, it seems that designed queues can further improve the model performances. It confirms that queues may be able to minimize the error term between the expected and the empirical ones. Comparing with variant 7, it seems that continuity constraint is effective.

\section{Analysis}

\begin{figure}[t]
\centering
\includegraphics[width=\columnwidth]{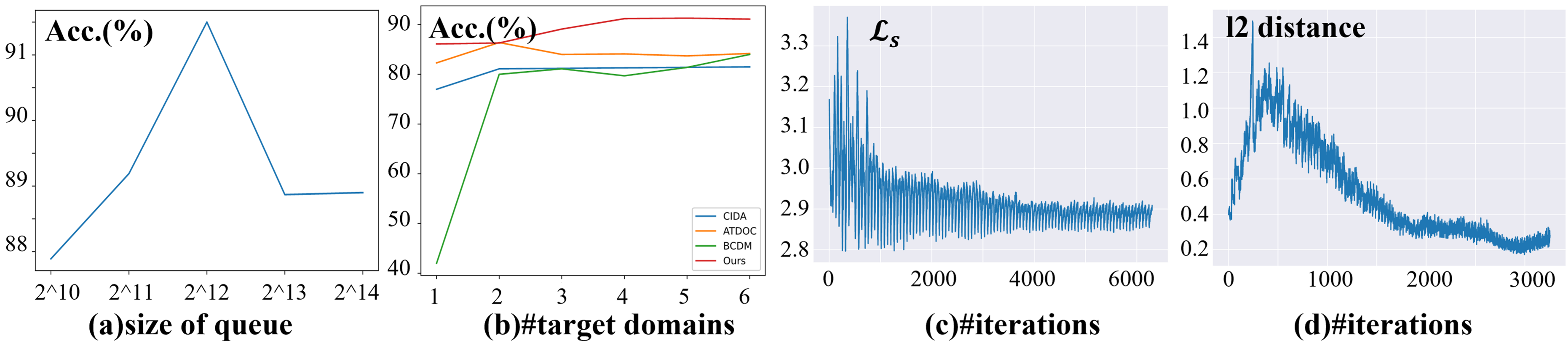}

\caption{(a) Analysis of queue size. (b)Analysis of the number of probe target domains. (c)$\mathcal{L}_s$. (d) $l$2 distance between two classifiers' predictions.}
\label{curve}
\end{figure}



\begin{figure}[t]
\centering
\includegraphics[width=0.95\columnwidth]{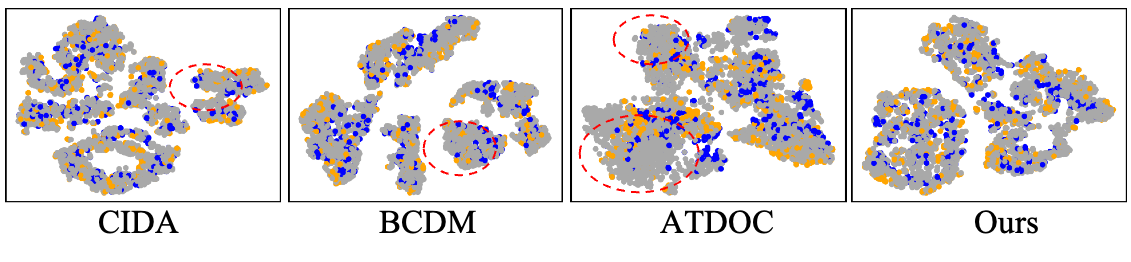}

\caption{Visualization of features obtained from encoder using t-SNE \cite{van2008visualizing}. Orange, blue and grey points indicate source1, source2 and target samples from the test set of SmallNorb respectively. Red circles denote the target domain dense distribution, i.e., large source-target discrepancy.}
\label{tsne}
\end{figure}





\subsection{Discussions}
\textbf{Why are some methods worse than SO?} It demonstrates that CDA is a challenging problem, and most existing DA methods can not be applied directly. The reason is twofold: 1) It may be because the adversarial-based methods only align the observed probe target domain and source domain but do not consider the unseen target domain. As a result, the model overfits on source and probe target domains. 2)These methods propose one classifier for all domains. Therefore, it is hard to fit all domains with large discrepancies. Meanwhile, it shows the advantage of proposing two classifiers. 

\textbf{The choice of two source domains its effects} We set up this novel benchmark to analyze this point. If two domains are similar, the adaptation performances degrade. It can be seen in Tables \ref{result} that P4 achieves the worst performance. 

\subsection{Hyper-parameter Analysis}
\textbf{Queue size.} We evaluate our methods with different queue sizes, and the results are visualized in Figure \ref{curve}(a). It can be observed that the accuracy curve raises at the beginning, and then falls. It is because that the queue enlarges the number of samples that compute the discrepancy, thus it promotes the adaptation. But large size makes the queues have more features extracted by the model before several iterations, which lead to imprecise discrepancy computation.   

\textbf{Number of probe target domains.} We evaluate the task using different number of probe target domains in the experiment in Figure \ref{curve}(b). It can be observed that with the increasing number of probe target domains, the overall accuracy increases for all methods. Our method always achieves the best-performing method for different setups. 

\subsection{Qualitative Analysis.} We perform t-SNE plot of the two source domain features and target features in Figure \ref{tsne}. The more evenly the points of three colors are mixed, the better the domains are aligned. We highlighted some areas using red dashed circles in the subfigure of CIDA, BCDM and ATDOC. In these areas, grey dots (target) dominate, with very few blue (source 1) and orange (source 2) dots overlapping with grey ones. However, this pattern does not appear in the t-SNE plot of our method, where dots of three colors are mixed more evenly. It seems that our method achieves the best domain alignment on unseen target domains as the source features spread more in the target zones.

\subsection{Convergence Analysis}In Figure \ref{curve}(c), we plot the loss $\mathcal{L}_{s}$ with respect to the number of iterations in the training procedure of smallNorb. The loss implies the distance between all target domains and the trajectory, that is -- the aim of this is to verify the effectiveness of our \textit{Pull Step}. It can be observed that our method achieves comparable convergence and relatively small loss, which suggests that target domains are able to be pulled close to the trajectory. 

In Figure \ref{curve}(d), we visualize the $l2$ distance between two classifiers' predictions of the target sample. The aim is to demonstrate the distance between two source domains in the trajectory, that is, verifying the effectiveness of our \textit{Shrinkage Step}. \textit{First}, due to discrepancies between two source domains and the high impact of cross-entropy loss in the beginning, two classifiers learn to discriminate in their individual domain. Therefore, the $l2$ distance starts increasing. Next, due to our proposed discrepancy measurement and reduction losses dominating the training, domains are aligned together and the two classifiers are pulled closely (distances are small). The convergence implies the shrinkage of the trajectory formed by all domains. 

Note that we can observe the regularly alternating patterns of the loss functions in Figure \ref{curve}(c-d). This is because the training procedure is in the alternating manner and the two steps are adversarial. The above two convergences jointly demonstrate that our method is able to reduce the discrepancy progressively.





\section{Conclusions}
In this work, we investigate a novel problem namely the continuous domain adaptation. We have proposed a novel framework for this CDA problem that outperforms other baseline techniques.

\begin{acks}
This work is supported by Peking University Medicine Seed Fund for Interdisciplinary Research (BMU2022MX011), the Fundamental Research Funds for the Central Universities and PKU-OPPO Innovation Fund (BO202103), and Zhe jiang Lab (NO.2022NB0AB05).
\end{acks}
\balance
\bibliographystyle{ACM-Reference-Format}
\bibliography{ref}

\appendix
\clearpage
In the following, we provide additional materials to support our submission (omitted from the main submission due to space limits). Specifically, we \textit{first} provide implementation details including benchmark, network architecture, training procedure and details of comparison methods. \textit{Then} we analyze the datasets used in CDA for more details. \textit{Next}, more comparisons are performed and ablation studies are provided, even including another evaluations using existing multi-domain adaptation benchmarks. This proves that our method is able to be extended to other benchmarks and achieves robust and best performing results. \textit{Finally}, we analyze, discuss and re-emphasize the motivations, main observations and insights in the new Continuous Domain Adaptation (CDA) task.

\section{Additional Implementation Details}

In this section, we provide additional information of the dataset, evaluation metrics and the implementation details of our and the comparison methods in the CDA task.  

\subsection{Datasets details}

\textbf{Rotating EMNIST}: \cite{cohen_afshar_tapson_schaik_2017} is an extension of MNIST to handwritten letters. It provides six different splits, and we use EMNIST(balanced) which contains 131,600 characters of 47 classes. Images are of resolution $32 \times 32$ and rotated by $0^{\circ}$ to $180^{\circ}$ 

\textbf{SmallNorb}: \cite{lecun2004learning} contains 48,600 images of 50 toys from 5 generic categories: four-legged animals, human figures, airplanes, trucks, and cars. The toys were imaged under 9 elevations (30 to 70 degrees every 5 degrees). Each image is downsampled to $48 \times 48$ pixels and cropped to a size of $32 \times 32$ pixels.
    
.  
    
\textbf{Age Face Synthesis}: Age is a common and continuous attribute that causes humans' appearance variations, however to the best of our knowledge, we have not found any existing dataset to support the study. Therefore, we propose to use Lifespan Age Transformation Synthesis \cite{orel2020lifespan} to synthesize a full span of 0–90 ages with the interval of 10. To be specific, 1000 identity images are selected from CelebA\cite{liu2015faceattributes} under synthesis based on the top 1,000 FID scores \cite{xu2018empirical}. Finally, the Face Age Synthesis contains 294,330 images of resolution $256 \times 256$ in total. 

\subsection{Experiment Split Choices}
    
First, we introduce four \textbf{configurations} based on the source domain locations in the attribute space denoted as P1-P4, as shown in Figure 6 in the main paper. In P1, where two source domains (red dots) are at the edge of the domain space, we randomly sample 50\% of the domains (purple dots) in the middle as probe target while others as unseen target test domains. As for P2-P4, two source domains divided the continuous domain into three segments, so we designed four settings S1-S4 based on which segment we shall select to randomly sample the probe target. In each selected segment we randomly sample 50\% as the probe target domain.
    
For the sake of better reproductive research, the 13 splits in SmallNorb and Age Face Synthesis are listed in Table \ref{spsmall}. All the numbers in the table indicate domain indexes. In SmallNorb, 0-8 refer to 9 elevations (30 to 70 degrees every 5 degrees). In Age Face Synthesis, 0-9 refer to 10 ages (0 to 90 degrees every 10 years). In Rotating EMNIST, we rotate a batch of images with different angles to generate source and probe target domain. The angles are listed in Table \ref{spemnist}.

\begin{table}[h]
\small
\caption{Splits in SmallNorb and the Face Age Synthesis.In SmallNorb, 0-8 refer to 9 elevations (30 to 70 degrees every 5 degrees). In Age Face Synthesis, 0-9 refer to 10 ages (0 to 90 degrees every 10 years).}
\begin{tabular}{c|cccc|c|cccc}
\hline
&\multicolumn{4}{c|}{SmallNorb} & &\multicolumn{4}{c}{Age  Face  Synthesis} \\
P &S1 &S2 &S3 &S4 &P &S1 &S2 &S3 &S4\\
\hline
 
0,8 & 2,4,5  & -- & --  & --  &0,9& 6,1,5& -- & -- & -- \\
2,6 & 1,5,8 &1,5 & 5&1 &1,7  &0,3,8 &3,8 &8 &3    \\
3,6& 0,5,8 &0,5 &5 &0  &2,6  & 1,5,8 &1,5 &5 &1   \\
4,6 & 0,5,8 &0,5 &5 &0 &3,6  & 0,5,8 &0,5 &5 &0 \\ 
\hline
\end{tabular}

 \label{spsmall}
\end{table}
\begin{table}[h]
\small
\caption{Splits in Rotating EMNIST.}
\begin{tabular}{c|cccc}
\hline
P&S1 &S2 &S3 &S4\\
\hline
0,180& (0,180) & -- & -- & --\\
22.5,157.5 & (0,180) &(0,157,5) &(22.5,157.5)&(0,22.5) \\
45,135   & (0,180) &(0,135)&(45,135)&(0,45)  \\
67.5,112.5  & (0,180) &(0,112.5)&(67.5,112.5)&(0,67.5)\\       
\hline
\end{tabular}
 \label{spemnist}
\end{table}

\begin{figure}[t]
\centering
\includegraphics[width=0.8\columnwidth]{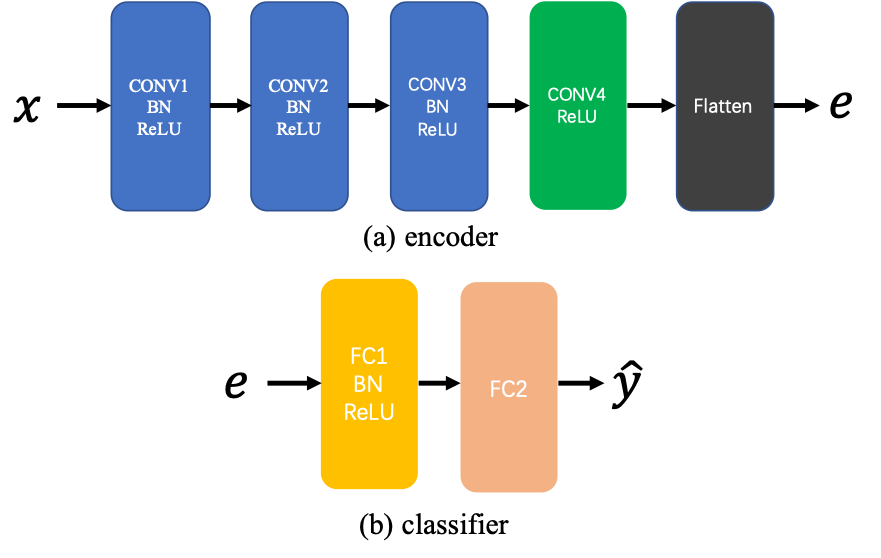} 
\caption{Network architecture and components. (a) Encoder; (b) Classifier.}
\label{module}
\end{figure}
\begin{table}[h]
\small
\caption{Network architecture of source-only model in SmallNorb and Rotating EMNIST. The output channels are 5 and 47 in SmallNorb and Rotating EMNIST, respectively.}
\begin{tabular}{c|cccc}
\hline
layer name&in channels &out channels &kernel size &stride  \\
 \hline
Conv1&1&256&3&2\\
Conv2&256&256&3&2\\  
Conv3&256&256&3&2\\  
Conv4&256&100&4&1\\ 
FC 1 &100 &256 & -- & --\\
FC 2 &256 &5/47 & -- & --\\
\hline
\end{tabular}

 \label{module-network-table}
\end{table}

\subsection{Network Architecture.}

The network architectures are shown in Figure \ref{module}. The network architecture used in SmallNorb and Rotating EMNIST is shown in Table \ref{module-network-table}. For Face Age Synthesis, we used the ResNet50 encoder part pretrained on ImageNet as the main encoder. We only change the classifier to the 1000 classes. 



\noindent\textbf{Comparison Method Details}

\begin{itemize}
    \item \textbf{Source-Only Model}: The source only model consists of an encoder and a classifier, as shown in Figure \ref{module} (a) and (b) respectively. In smallNorb and EMNIST, the encoder and classifier design are shown in Table \ref{module}. In Face Age Synthesis dataset, the encoder follows the ResNet50 network but change the final layer classifer in our application. 
    
    \item \textbf{CIDA}: In the CDA formulation, it is assumed that the domain indexes are unavailable. However the CIDA method needs domain index to work so we assign pseudo domain index for each domain in the experiment. Specifically, we assign 0 and 1 to source domain 1 and 2 respectively and all probe target domains are assigned with 0.5. For fair comparisons, we used the same encoder as source-only model in all experiments.
    
    \item \textbf{BCDM, ATDOC}: BCDM and ATDOC aim to adapt on single source domain. Thus, we combine two source domains. For fair comparisons, we used the same encoder, classifier as the source-only model in all experiments.
    
\end{itemize}

\end{document}